\documentclass[11pt]{article}

\usepackage{graphicx}
\usepackage{amsmath}
\usepackage{booktabs}
\usepackage{hyperref}
\usepackage{tabularx}
\usepackage{float}
\usepackage{amsmath}
\usepackage{float}
\usepackage[section]{placeins}
\usepackage{cite}

\title{A Hybrid Machine Learning Model for Cerebral Palsy Detection\thanks{This paper is based on a previously published article in the \textit{International Journal of Intelligent Systems and Applications in Engineering} (IJISAE), 2024. Available at: https://ijisae.org/index.php/IJISAE/article/view/7390}}

\author{
Karan Kumar Singh \\
Department of Computer Science and Engineering\\
Sharda University, India\\
\texttt{karankumarsingh7870@gmail.com}
\and
Nikita Gajbhiye \\
Department of Computer Science and Engineering\\
Sharda University, India
\and
Gouri Sankar Mishra \\
Department of Computer Science and Engineering\\
Sharda University, India
}

\begin{document}

\maketitle

\begin{abstract}
The development of effective treatments for Cerebral Palsy (CP) can begin with the early identification of affected children while they are still in the early stages of the disorder. Pathological issues in the brain can be better diagnosed with the use of one of many medical imaging techniques. Magnetic Resonance Imaging (MRI) has revolutionized medical imaging with its unparalleled image resolution. A unique Machine Learning (ML) model that was built to identify CP disorder is presented in this paper. The model is intended to assist in the early diagnosis of CP in newborns. In this study, the brain MRI images dataset was first collected, and then the preprocessing techniques were applied to this dataset to make it ready for use in the proposed model. Following this, the proposed model was constructed by combining three CNN models, specifically VGG 19, Efficient-Net, and the ResNet50 model, to extract features from the image. Following this, a Bi-LSTM was utilized as a classifier to determine whether or not CP was present, and finally, the proposed model was employed for training and testing. The results show that the proposed model achieved an accuracy of 98.83\%, which is higher than VGG-19 (96.79\%), Efficient-Net (97.29\%), and VGG-16 (97.50\%).. When the suggested model is compared to other models that have been pre-trained in the past, the accuracy scores seem to be much higher.\\


\end{abstract}

\section{Introduction}

Cerebral palsy (CP) is a collection of impairments affecting mobility, posture, and motor function. These problems are caused by non-progressive injuries to the immature brain [1]. Early identification of CP or high-risk CP in childhood is important so that targeted therapies and support services can be provided promptly.. There are two separate ways that CP can be detected early, according to guidelines published in 2017 by Novak et al., [2]. One way is for babies who have newborn-detectable risk factors, like being born prematurely, having a low birth weight, or having hypoxic-ischemic encephalopathy, to be checked for CP before they are 5 months old. This can be done with a thorough history and physical examination, as well as with neonatal MRI, General Movement Assessment (GMA), and the Hammersmith Infant Neurological Examination (HINE) [3]. Second, between the ages of 5 and 24 months, infants who do not have any known neonatal risk factors but do have infant-detectable risk factors, such as delayed motor milestones during the infantile period, should be evaluated for CP using a battery of tests, including standardized motor assessments, the HINE, and neonatal MRI [4,5].
Clinical assessment of symptoms and disorders remains the primary method for CP diagnosis (Table 1). The more quickly these patients get a proper diagnosis of CP and begin treatment, the higher their chances of a full recovery with fewer long-term effects on their lives. Figure 1 displays the many forms of CP. In the end, CP is a complex disorder with several risk factors, the majority of which are not easily preventable. When it comes to pediatric CP, the worldwide agreement among industry professionals is that the best course of action is early diagnosis and treatment. Children have a higher chance of making a full recovery if they get early interventional therapy, which can significantly slow the progression of their illness [6,7].

\begin{table}[htbp]
\centering
\caption{Common genetic and metabolic diseases that behave similarly to CP}
\renewcommand{\arraystretch}{1.3}
\begin{tabular}{|p{5cm}|p{5cm}|p{5cm}|}
\hline

\textbf{Disorders with prominent spasticity} &
\textbf{Disorders with prominent dyskinesia} &
\textbf{Disorders with prominent ataxia} \\

\hline

\begin{itemize}
\item Hereditary spastic paraplegias
\item Arginase deficiency
\item COL4A1-related spastic CP
\item Biotinidase deficiency
\item Aicardi-Goutières syndrome
\item Sulfite oxidase deficiency / Molybdenum cofactor deficiency 
\item Leukodystrophies such as metachromatic leukodystrophy, adrenoleukodystrophy, Sjogren-Larsson syndrome 
\end{itemize}

&

\begin{itemize}
\item Dopa-responsive dystonia
\item Sepiapterin reductase deficiency
\item Glutaric aciduria type 1
\item Glucose transporter deficiency type 1
\item Neurodegeneration with brain iron accumulation
\item Cerebral creatine deficiency syndrome
\item Lesch-Nyhan syndrome
\item Cerebral folate deficiency
\item ADCY5-related dyskinesia
\item PCDH12-related dyskinesia 
\item NKX2-1 related ataxic dyskinetic CP 
\item TSEN54 gene-related pontocerebellar hypoplasia type 2 
\end{itemize}

&

\begin{itemize}
\item Glucose transporter deficiency type 1
\item Ataxia telangiectasia
\item Pelizaeus-Merzbacher disease
\item Hereditary ataxias
\item Joubert syndrome
\item Mitochondrial cytopathies (mainly 8993 mutation) 
\item Pontocerebellar hypoplasia
\item Cockayne syndrome 
\item Niemann-Pick disease type C
\item Angelman syndrome
\item Gangliosidosis type 1 (juvenile and adult forms)
\item Non-ketotic hyperglycinemia
\item Maple syrup urine disease
\item NKX2-1 related ataxic dyskinetic CP
\end{itemize}

\\
\hline

\end{tabular}
\end{table}

\begin{figure}[h]
\centering
\includegraphics[width=0.7\linewidth]{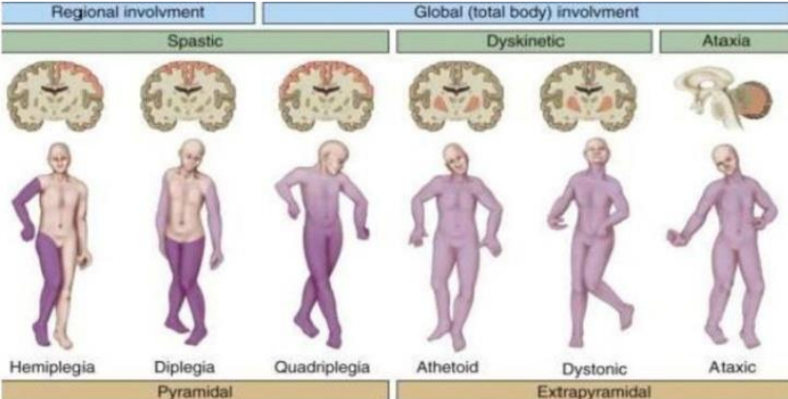}
\caption{Types of CP disorders}
\end{figure}

Medical Image Processing is a single illustration of how recent advances in AI and ML have revolutionized the medical field. This technique has made disease diagnosis much easier and faster than in the past, when the process was difficult and time-consuming [20]. Consequently, this research presents a machine-learning model that can identify CP diseases by analyzing large databases of patient records and brain imaging data \cite{ref54}. Therefore, the goal of using brain MRI images is to discover a way to diagnose CP that combines a data augmentation methodology with a Transfer Learning (TL) model [21]. The goal of the model is to help doctors make better judgments for their patients by incorporating ML into diagnosis. After that, the remaining parts of the paper are organized as follows: In section 2, a literature review of the research problem is presented. The explanation of the issue is presented in Section 3, and the research objectives in section 4. The formulation of the proposed model and technique is discussed in Section 5. Following the presentation of the findings and analysis of the research in Section 6, the discussion of the study is presented in Section 7. This research subject is brought to a conclusion in section 8.

\section{Literature Review}

In this section, the authors study some previous work and find some future gaps based on CP Disorder Detection using ML. Table 2 shows the comparison of previous work. 

\begin{table}[H]
\centering
\caption{Comparison of previous work}
\small
\renewcommand{\arraystretch}{1.2}
\setlength{\tabcolsep}{4pt}

\begin{tabularx}{\textwidth}{|X|X|X|X|X|X|}
\hline

\textbf{Authors (Year) [Reference]} &
\textbf{Participants} &
\textbf{Dataset collected} &
\textbf{Techniques Used} &
\textbf{Outcomes} &
\textbf{Research Gap} \\

\hline

Mathew et al., (2024) [22] &
10 participants (7 males and 3 females) &
Holland Bloor View Kids Rehabilitation Hospital &
RF &
The RF classifier performed better in the per-user validation scenario with an F1 score of 0.896 (SD 0.043). &
Future research should collect data in less controlled real-world conditions to evaluate performance changes. \\

\hline

Sabater et al., (2024) [23] &
53 participants (19 females and 34 males) &
Balearic Islands (Majorca, Spain) and Toledo (Castilla-La Mancha, Spain) &
InceptionV3 &
The suggested model achieved 62.67\% accuracy and an F1 score of 61.12\%. &
Highlights the necessity of further research and technological adaptation in healthcare applications. \\

\hline

Mohan et al., (2024) [24] &
19 patients &
--- &
Artificial Intelligence Assisted Learning Methodology (AIALM) &
The model achieved 96.4\% accuracy in predicting CP from brain scans. &
Accuracy confirmed through clinical evaluation and cross-validation. \\

\hline

Bertoncelli et al., (2023) [25] &
486 children with CP &
Lanval University Children’s Hospital and Nice Day Hospital &
Logistic Regression (LR) &
The multivariate model achieved 84\% accuracy with good sensitivity and specificity. &
Future work aims to improve the model using larger datasets. \\

\hline

Gao et al., (2023) [26] &
1204 patients &
Shanghai Children’s Hospital, Shanghai Jiao Tong University &
DL-based Motor Assessment Model (MAM) &
External validation shows strong performance with AUC = 0.967. &
ML-based GMA automation could improve early CP screening worldwide. \\

\hline

Ramadhan et al., (2022) [27] &
98 normal MRI images and 65 CP MRI images &
Brain CP MRI dataset &
CNN &
The model achieved perfect sensitivity, accuracy, precision and F1-score. &
Future applications include CP subtype classification. \\

\hline

Li et al., (2022) [28] &
5514 records from 1755 patients &
Chinese Academy of Traditional Chinese Medicine dataset &
Knowledge-Based RNN (KBRNN) &
Diagnostic accuracy improved from 79.31\% to 83.12\%. &
Future work should train with larger EMR datasets. \\

\hline

Cheragh et al., (2022) [29] &
21 CT scans of children with CP &
Hip image dataset &
U-Net with Transfer Learning &
IoU = 0.99, recall = 0.98 and Dice score = 0.99. &
Task focuses only on segmentation without additional diagnostic options. \\

\hline

Xue et al., (2021) [30] &
26 participants &
St. Louis Children's Hospital &
Weighted Logistic Regression (WLR) &
Accuracy = 86–89\%, sensitivity = 88–92\%, specificity = 70–75\%. &
Future research should improve sleep pattern detection. \\

\hline

Zhu et al., (2021) [31] &
--- &
MINIRGBD dataset &
Deep Learning &
System achieved 91.67\% accuracy outperforming other DL models. &
Further research needed for improved visualization techniques. \\

\hline

\end{tabularx}
\end{table}

\section{Problem Statement}

Detecting CP in newborns is challenging because the symptoms may not become apparent until approximately one year after birth. This results in a reduction in the likelihood of achieving better outcomes from therapy when compared to the situation in which the condition was diagnosed earlier. However, conventional diagnostic approaches often depend on individual assessments and can be time-consuming, which can result in delays in both the diagnosis and the treatment of the condition. \\

The proposed solution involves developing a machine-learning model that can analyze medical data and accurately detect the presence of CP. The model aims to leverage advanced algorithms to identify patterns and indicators of CP in various types of medical data. By automating the detection process, the model seeks to reduce diagnostic errors, expedite the diagnosis, and ultimately improve patient outcomes by enabling earlier interventions. This approach not only aims to enhance diagnostic accuracy but also to provide a scalable and accessible tool for healthcare providers worldwide.

\section{Research Objective}

Here are the research objectives of the paper: \\
•	Conduct a comprehensive literature review to understand the current state of ML applications in predicting CP. \\
•	To develop and validate a ML-based prediction model specifically designed to identify the risk of CP disorder in early childhood.\\
•	To compare measures including accuracy, precision, recall, F1-score, and area under the receiver operating characteristic (ROC) curve to assess the efficacy of different ML algorithms in predicting CP.\\
•	To evaluate the suggested prediction model in comparison to current models and methods, drawing attention to its strengths and areas for improvement.

\section{Research Methodology}

In this section, the author provides the research methodology flowchart based on the study and design of an optimized prediction model for CP disorder using ML techniques (Figure 2). The first step involves gathering comprehensive data on patients diagnosed with CP. This data can be collected from hospitals, clinical trials, and healthcare databases. The dataset includes several attributes such as demographics, medical history, genetics, clinical test results, and other relevant health indicators. Cleanup and preprocessing are necessary steps after data collection to make sure the data is ready for analysis. As part of this process, we address missing values, standardize numerical data, encode categorical variables, and eliminate outliers that can distort the result. Feature extraction refers to the steps used to turn unstructured data into trainable model features. In this stage, we will determine which factors are most important for CP prediction. For feature extraction, they used a TL model, such as VGG-19, Efficient-Net, or VGG-16. The next step is to divide the prepared dataset into two primary sets: one for training purposes and another for testing. A 50:50 split is the norm for data distribution. The current step is to find the optimal model for CP prediction using ML techniques, such as Bi-LSTM using TL approaches. Based on the evaluation metrics, they evaluate the efficiency of various models. The best model for predicting CP can be found by comparing these models based on evaluation metrics.

\begin{figure}[h]
\centering
\includegraphics[width=0.7\linewidth]{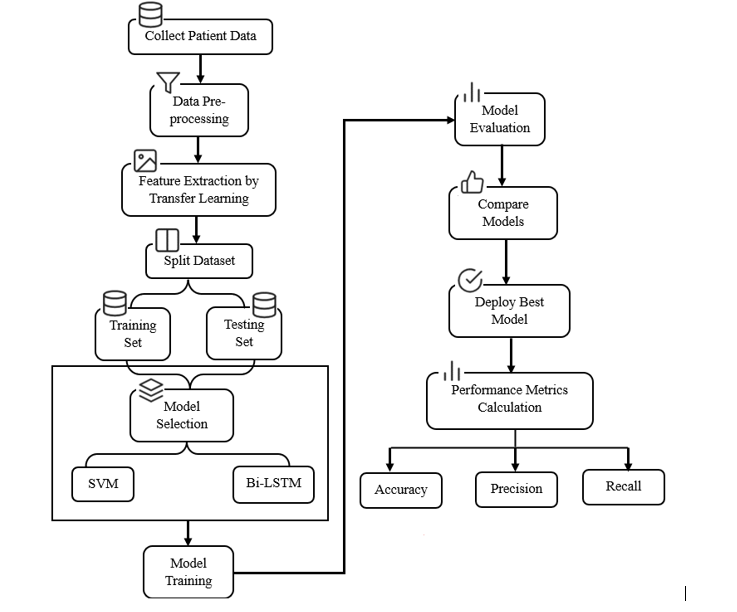}
\caption{Flowchart of proposed work}
\end{figure}

5.1	Data Collection \\
This phase included the collection of data that was used for training and testing the suggested model. It was necessary to have two distinct classes of MRI images of the brain: one class was for MRI images of the brain for normal individuals, and the other class was for MRI images of the brain for those who had CP .Due to privacy and ethical considerations, the MRI images collected from the hospital cannot be publicly shared. As for the images of the brains of individuals who have CP, there was no data set available on the internet, therefore they were taken from Santi Hospital in Agra from persons who have CP. There are a total of 65 photographs in this collection. There are examples of normal MRI brain images as well as MRI images of CP on display in Figure 3.

\begin{figure}[h]
\centering
\includegraphics[width=0.7\linewidth]{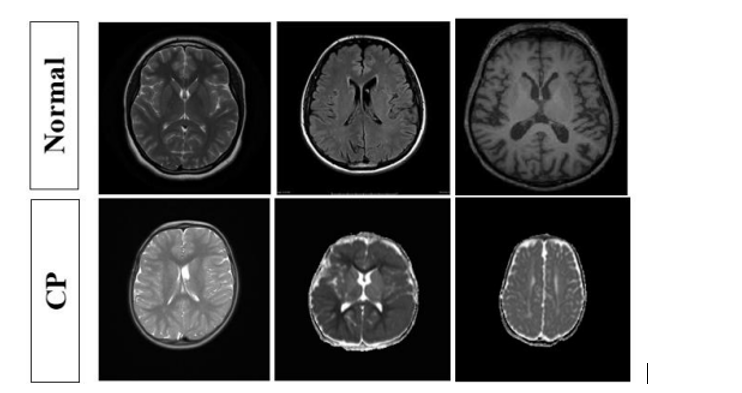}
\caption{MRI brain dataset}
\end{figure}

5.2	Data Pre-processing 

The data are being prepared for the categorization model at this stage of the process. This stage is comprised of two stages, which are the splitting of the data and the augmentation of the training data phases. \\

5.2.1	Data Splitting 

The dataset is now divided into two sections, one for training and the other for testing. The model is trained using the training dataset, and its efficacy is assessed using the testing dataset \cite{ref56}. When using ML, it is common practice to use a dataset that was not used during training to test the model. The dataset plays a vital role in training and evaluating the model. The MRI images are the focus of this study. The MRI images in the dataset are of two types: one is for individuals with CP (abnormal), and the other is for those without C (normal). They used Kaggle to get both kinds of MRI images. One type is used for training and the other for testing, and each sort is split into two groups. Part of this research included dividing the dataset in half so that 50 percentage could be used for testing and 50 percentage for training.\\

5.2.2	Data Augmentation (DA) 

Data augmentation is a method for improving the quality of data included in a dataset. It is also used to create new pictures that are derived from existing ones. DA is now the default in ML due to its ease of use and the fact that ML's model training approach demands a large amount of data for good model training. This study makes use of both rotational and flipping augmentation techniques. Figure 4 (a) shows the original image with the DA applied to it; Figure 4 (b) shows the rotation process DA technique, and Figure 4 (c) shows the flipping process DA method.

\begin{figure}[h]
\centering
\includegraphics[width=0.7\linewidth]{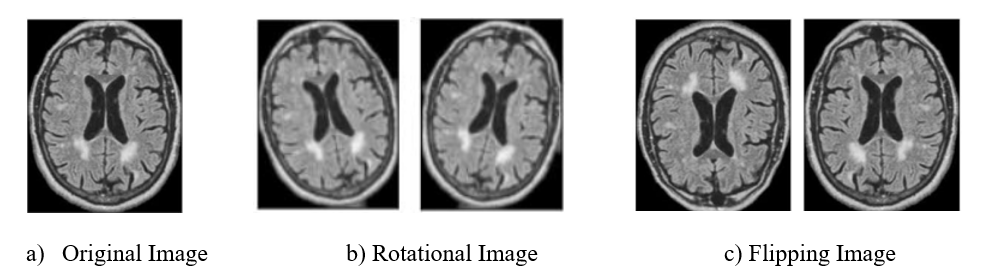}
\caption{Data Augmentation}
\end{figure}

5.3	Transfer Learning (TL)

TL is an essential and effective method for training a network with a small amount of data [32]. In this study, they use VGG-16, VGG-19, and Efficient-Net model for CP detection.\\

5.3.1	Efficient-Net 

In 2019, a family of deep CNN architectures called Efficient-Net was released. After that, it has shown breakthrough performance on several computer vision applications. With a compound scaling strategy, Efficient-Net optimizes the network's depth, breadth, and resolution all at once, resulting in great accuracy and computing efficiency. The two main networks of Efficient-Net are the backbone network, which does feature extraction from input images, and the head network, which does the final classification [33]. To capture input spatial and channel-wise correlations, the backbone network includes a mix of mobile inverted bottleneck convolutional layers and squeeze-and-excitation (SE) blocks\cite{ref55}. For the last categorization, the head network employs fully linked layers in conjunction with global average pooling. Figure 5 shows the framework of the Efficient-Net method.\\

\begin{figure}[h]
\centering
\includegraphics[width=0.7\linewidth]{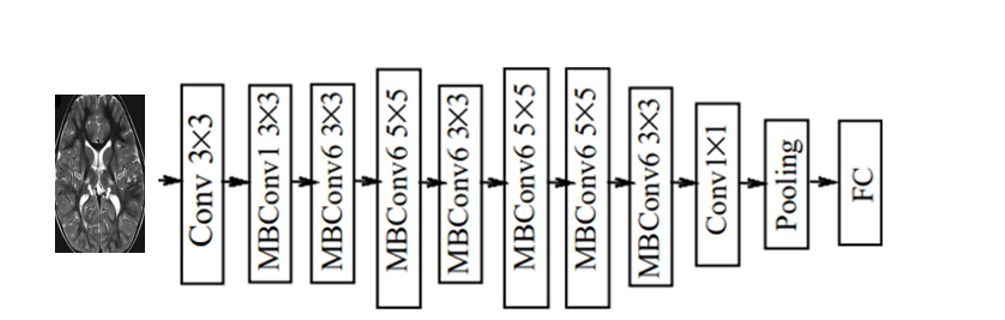}
\caption{Framework of Efficient-Net method [34].}
\end{figure}

5.3.2	VGG-19 Model

The VGG-19 neural network is a traditional convolutional neural network that can extract high-level features that are more complex [35]. A total of 16 convolutional layers and 3 fully connected layers makes up its network architecture, where convolutional layers could be represented as five blocks, each comprising two to four convolutional layers of varying sizes [36,54]. The VGG-19 model extracts features from one or more network layers to create high-level feature layers in Tub-sGAN that describe image style and content [37,38]. The structure of the VGG-19 model is seen in Figure 6.\\

\begin{figure}[h]
\centering
\includegraphics[width=0.7\linewidth]{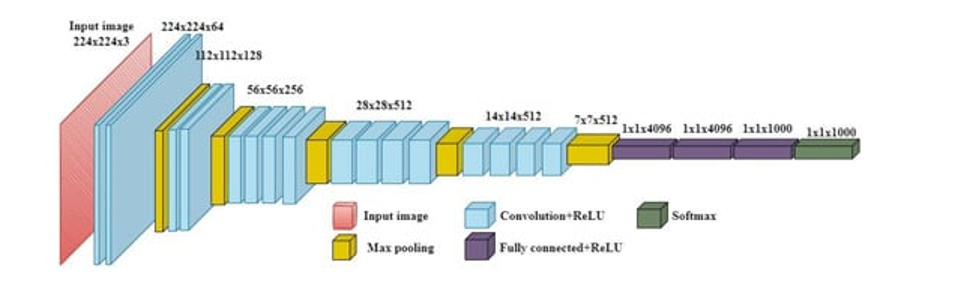}
\caption{Representation of the VGG-19 architecture [39].}
\end{figure}

5.4	Machine Learning (ML)

 ML techniques are a set of statistical procedures used to classify and organize attitudes according to certain attributes. In addition to training on several instances, these methods provide a roadmap for classifying new cases. In this method, the best match is used to choose the optimal combination that can provide new data. Using the training data classifications as a foundation, this ML technique was trained using a small set of predefined categories and then trained on new data [40]. Figure 7 illustrates the ML techniques. \\

\begin{figure}[h]
\centering
\includegraphics[width=0.7\linewidth]{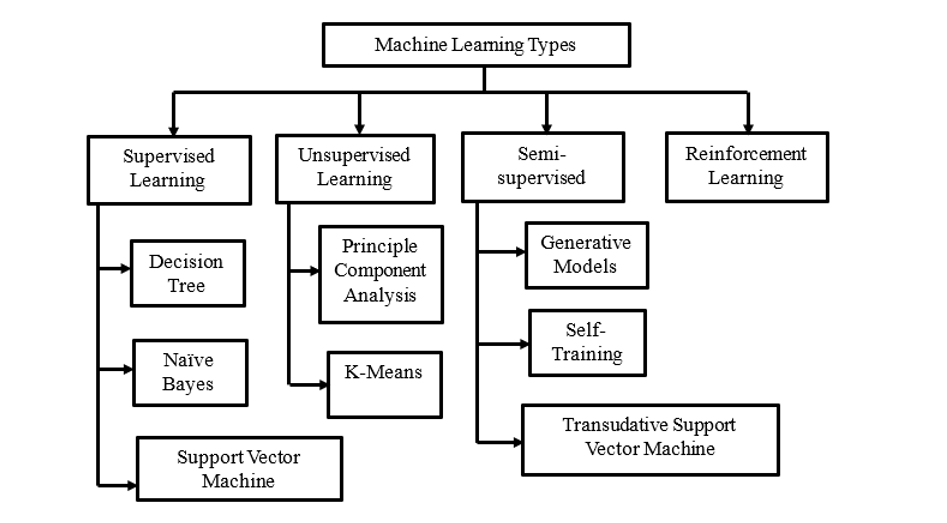}
\caption{Various types of ML techniques [41-43].}
\end{figure}
 
5.4.1	Bi-LSTM 

Bi-LSTM is a developed architecture from regular LSTM that could be used to improve sequence classification difficulties [44]. As shown in Figure 8, an LSTM cell consists of three essential components: an input gate, a forget gate, and an output gate. One of the main roles of input gates is to control the flow of fresh data into memory. The forget gate is responsible for storing items in memory for a certain duration. The quantity of memory needed to activate the block is finally controlled by the output gate [45,46]. The ability for networks to acquire forward and backward sequence information at each time step is made possible by Bi-LSTM architecture. One way in which Bi-LSTM varies from its traditional version is its capacity to train input using two distinct techniques. Two methods are available for preserving information: one that goes back in time and another that goes forward in time\cite{ref53}. Using the two hidden states together, it is feasible to store information from both the past and the future, and there's a chance to store information from the future as well.\\

\begin{figure}[h]
\centering
\includegraphics[width=0.7\linewidth]{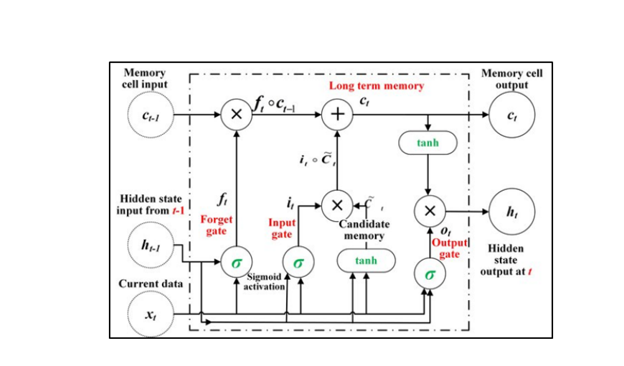}
\caption{An LSTM network [47,48].}
\end{figure}

5.5	Performance Metrics Calculation

To evaluate the performance of the proposed classifier, four evaluation metrics were used, all based on the number of True positives (TP): This indicates that a person has been categorized as having CP; In a true negative (TN) test, a healthy individual is considered to be in good health. In the case of a false positive (FP), a healthy individual is mistakenly identified as having CP. In the case of a false negative (FN), an individual with CP is mistakenly identified as being in good health. Table 3 shows the confusion matrices of binary classification.\\

\begin{table}[H]
\centering
\caption{Confusion matrices}
\renewcommand{\arraystretch}{1.4}

\begin{tabular}{|c|c|c|}
\hline
 & \textbf{Actual positive} & \textbf{Actual negative} \\ 
\hline
\textbf{Predicted positive} & True Positive (TP) & False Positive (FP) \\ 
\hline
\textbf{Predicted negative} & False Negative (FN) & True Negative (TN) \\ 
\hline
\end{tabular}

\end{table}

\begin{itemize}
\item \textbf{Accuracy:} Presents evidence of the effectiveness of the categorization system.
\end{itemize}

\begin{equation}
ACC = \frac{TP + TN}{TP + TN + FP + FN}
\end{equation}

\begin{itemize}
\item \textbf{Recall} is the ratio of successfully anticipated positive observations to the total number of positive observations in the training session.
\end{itemize}

\begin{equation}
SEN = \frac{TP}{TP + FN}
\end{equation}

\begin{itemize}
\item Specifically, the \textbf{F1-score} is the average of the recall and precision parameters.
\end{itemize}

\begin{equation}
F1\text{-score} = 2 \times \frac{Precision \times Recall}{Precision + Recall}
\end{equation}

\begin{itemize}
\item \textbf{Precision} refers to the degree of effectiveness with which the model provides an accurate diagnostic.
\end{itemize}

\begin{equation}
PRE = \frac{TP}{TP + FP}
\end{equation}

\section{Results and Analysis}

The data set that was used in the process of putting the suggested model into action and evaluating its effectiveness is offered in this paper. The model is divided into two parts: the first part deals with the results of the suggested model's training phase, where the training data is used, and the second part deals with the results of the evaluation phase, where the test data set is used to test the performance of the proposal method that was trained with the training data. \\

6.1	VGG-19

The batch size was set to 32, the learning rate was set to 0.001, the optimizer was set to Adagrade, and lastly, categorical cross-entropy was chosen as a loss function. They were able to achieve the best set of model training parameters. They trained and verified the VGG-19 model for a total of 50 epochs, and the graphs that represent the accuracy of the model are shown in Figure 9.

\begin{figure}[h]
\centering
\includegraphics[width=0.7\linewidth]{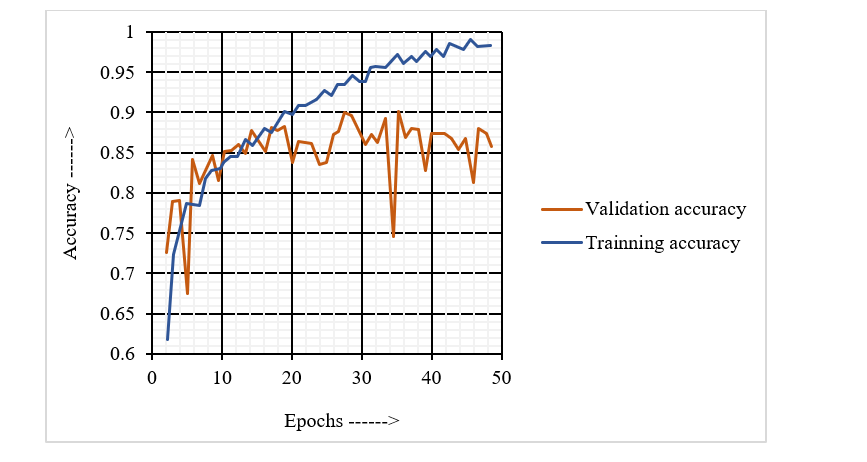}
\caption{Accuracy of the VGG-19 model}
\end{figure}

A representation of the VGG-19 model loss at each epoch is shown in Figure 10.

\begin{figure}[h]
\centering
\includegraphics[width=0.7\linewidth]{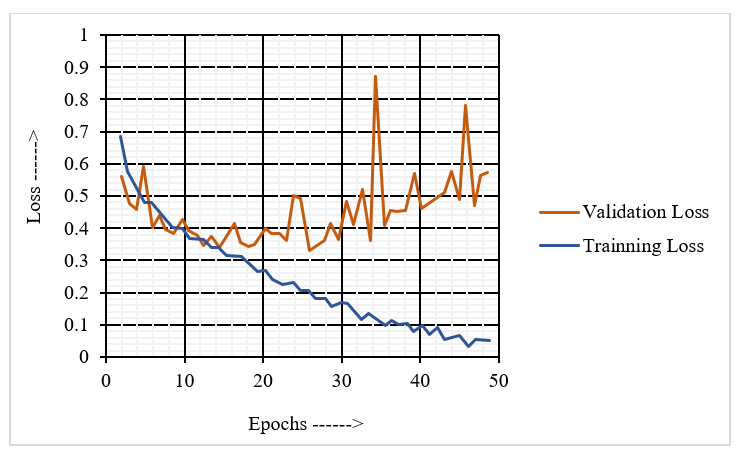}
\caption{ Loss of the VGG-19 model}
\end{figure}

Figure 11 illustrates the confusion matrix that is generated by using the VGG-19 model. The matrix has the following elements: TP = 19, FP = 1, TN = 19, and FN = 1.

\begin{figure}[h]
\centering
\includegraphics[width=0.7\linewidth]{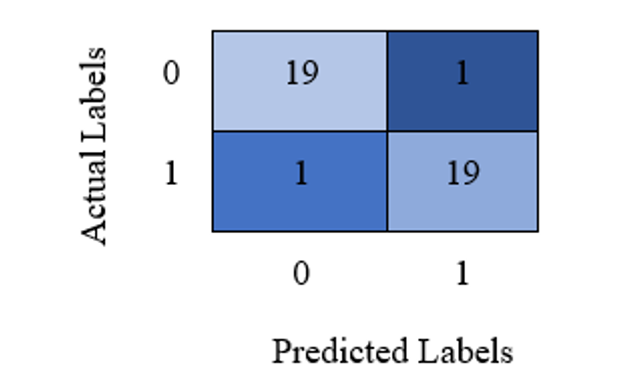}
\caption{Confusion Metrics of VGG-19 model}
\end{figure}

So, 19 of 20 individuals were thought to have CP, whereas 19 of 20 were thought to be normal. There is a report on the categorization using Efficient-Net in Table 6.

\begin{table}[H]
\centering
\caption{The result of VGG-19 model}
\renewcommand{\arraystretch}{1.3}

\begin{tabular}{|c|c|c|}
\hline
\textbf{S.No.} & \textbf{Performance Metrics} & \textbf{Result} \\
\hline
1 & \textbf{Accuracy} & 97.50\% \\
\hline
2 & \textbf{Precision} & 95.25\% \\
\hline
3 & \textbf{Recall} & 100\% \\
\hline
4 & \textbf{F1-score} & 97.56\% \\
\hline
\end{tabular}

\end{table}

6.2	Efficient-Net Model
Figure 12 shows the results of efficient-Net's validation and training/learning accuracies. The training accuracy, shown by the blue line, grows as the epoch count rises and reaches 100 percentage  after 50 epochs. The validation accuracy is shown by the brown curve, which starts at 96.56 percentage and rises to 98.02 percentage  after 50 epochs. Training was stopped after 50 epochs to avoid overfitting. Optimal training/validation accuracy was achieved by fine-tuning the number of training epochs.

\begin{figure}[h]
\centering
\includegraphics[width=0.7\linewidth]{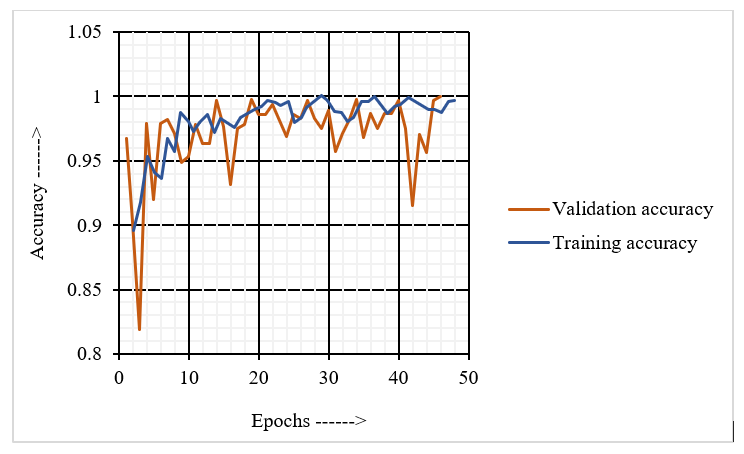}
\caption{ Accuracy of the Efficient-Net model}
\end{figure}

The training and validation loss for the Efficient-Net model is shown in Figure 13. A score of 0.0 would imply that the whole learning process was outstanding and there were no errors observed. After 50 epochs, the training loss reached 0.0030 and the validation loss, which started at 0.110 and decreased to 0.01, both decreased steadily with increasing epoch count.

\FloatBarrier

\begin{figure}[htbp]
\centering
\includegraphics[width=0.7\linewidth]{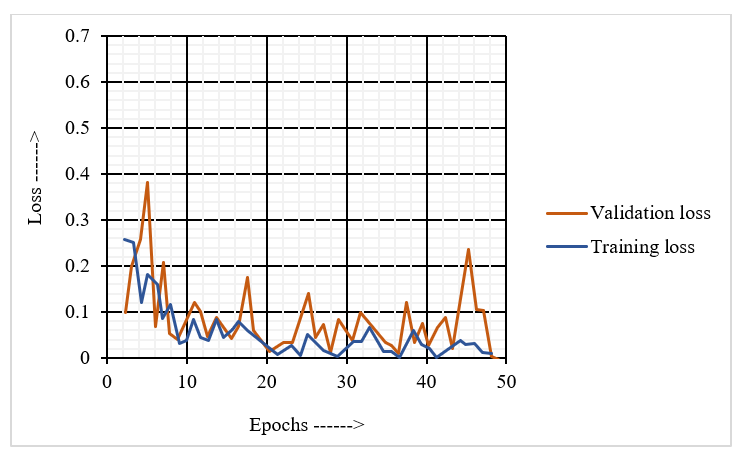}
\caption{ Loss curves of the Efficient-Net model}
\end{figure} 

Figure 14 illustrates the confusion matrix that is generated by using the Efficient-Net model. The matrix has the following elements: TP = 18, FP = 1, TN = 19, and FN = 2.

\FloatBarrier
\begin{figure}[htbp]
\centering
\includegraphics[width=0.7\linewidth]{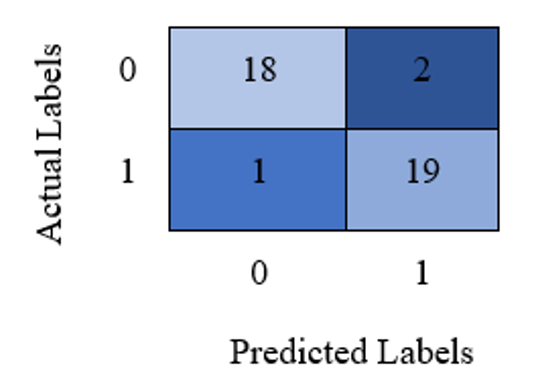}
\caption{ Confusion Metrics of Efficient-Net model}
\end{figure}

So, 18 of 20 individuals were thought to have CP, whereas 19 of 20 were thought to be normal. There is a report on the categorization using Efficient-Net in Table 5.

\begin{table}[H]
\centering
\caption{The result of the Efficient-Net model}
\renewcommand{\arraystretch}{1.3}

\begin{tabular}{|c|c|c|}
\hline
\textbf{S.No.} & \textbf{Performance Metrics} & \textbf{Result} \\
\hline
1 & \textbf{Accuracy} & 97.29\% \\
\hline
2 & \textbf{Precision} & 94.36\% \\
\hline
3 & \textbf{Recall} & 97.29\% \\
\hline
4 & \textbf{F1-score} & 95.80\% \\
\hline
\end{tabular}

\end{table}

6.3	The Proposed Model

The proposed model was implemented using VGG-19 and Efficient-Net architectures. This meant that every MRI image that was included in the training data set was processed by each of the available models. The results of an Efficient-Net model were 513 feature maps for each image, and the results of a VGG 19 model were 2062 feature maps. Following the integration of the feature map for every model, a 3047 feature map is created for every image.\\

6.3.1	Result of Training and Testing

The suggested model (VGG 19 and Efficient-Net) was built using weights from training each model on the training data set and the ImageNet dataset. Adam, the optimizer, teaches the model with a 0.4 learning rate and a hinge loss function. To make the model train more accurate, the training information goes through 50 iterations. Next, the model was tested using the MRI testing dataset, which includes 32 MRI images. Figure 15 shows the accuracy curve, whereas Figure 16 shows the loss curve. Because the validation curve is continually rising and falling while the training loss is constantly reducing, the training and validation curves show that the model started to overfit beyond a certain point. Due to the limited size of the training and validation sets, the model is unable to extract features for all possible scenarios.

\FloatBarrier
\begin{figure}[htbp]
\centering
\includegraphics[width=0.7\linewidth]{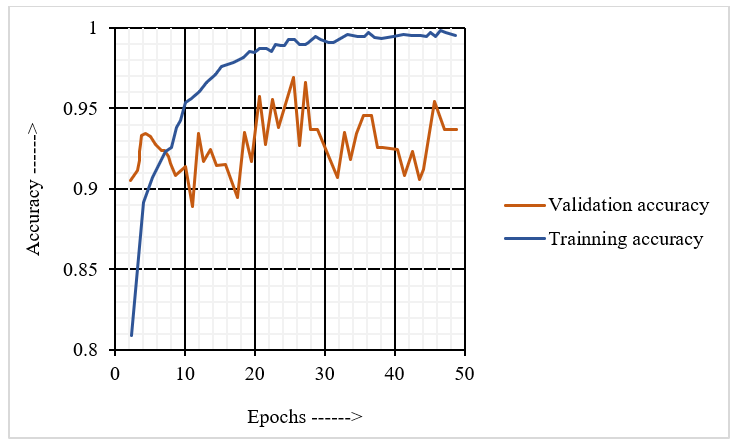}
\caption{ Accuracy curves of the suggested model}
\end{figure} 

\FloatBarrier
\begin{figure}[htbp]
\centering
\includegraphics[width=0.7\linewidth]{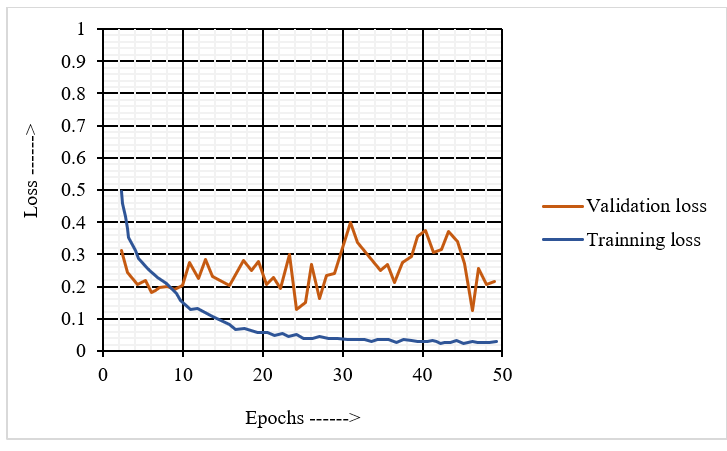}
\caption{ Loss curves of the suggested model}
\end{figure}

Additionally, the data that was used for this matrix matched the data that was tested. The confusion matrix provides a visual representation of the method's efficacy (Figure 17). Table 7 displays the classification report that was generated by utilizing the suggested model. The confusion matrix of the proposed model has the following values: TP = 20, FP = 0, TN = 19, and FN = 1. Twenty out of twenty persons were proposed to be normal, whereas nineteen out of twenty were proposed to be CP.

\FloatBarrier
\begin{figure}[htbp]
\centering
\includegraphics[width=0.7\linewidth]{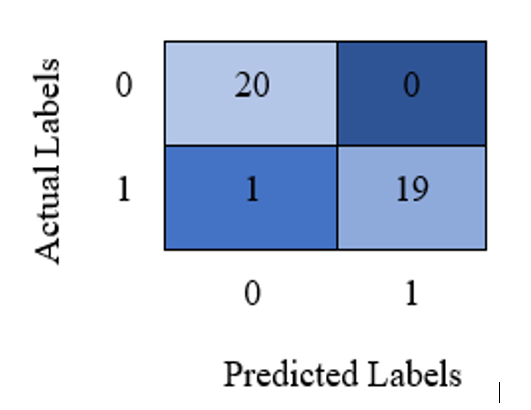}
\caption{ Confusion matrices of the suggested model}
\end{figure}

\begin{table}[H]
\centering
\caption{The result of the suggested model}
\renewcommand{\arraystretch}{1.3}

\begin{tabular}{|c|c|c|}
\hline
\textbf{S.No.} & \textbf{Performance Metrics} & \textbf{Result} \\
\hline
1 & \textbf{Accuracy} & 98.83\% \\
\hline
2 & \textbf{Precision} & 97.70\% \\
\hline
3 & \textbf{Recall} & 98.64\% \\
\hline
4 & \textbf{F1-score} & 98.17\% \\
\hline
\end{tabular}

\end{table}

6.4	Comparison analysis

A comparative analysis is conducted in this section, which is divided into two parts. The first part is a comparison of the proposed model with VGG-19 and Efficient-Net. In the second part, a comparative analysis of the suggested model with the approaches done in the past.\\

6.4.1	Comparison of the Suggested Model with VGG-19 and Efficient-Net

To predict CP, the VGG-19 and Efficient-Net models were utilized with a similar dataset that was used in the suggested model. The results of these models were compared with the findings of the proposed model. The accuracy that is acquired with the use of the VGG-19 model through the utilization of the TL technique is 96.79 percentage. The accuracy that is acquired with the use of the Efficient-Net model through the utilization of the TL technique is 97.29 percentage. The accuracy achieved by the VGG-19 model using transfer learning is 96.79\%. When applied to the same dataset, the suggested model attained an accuracy of 98.83 percentage. 

Figure 18 is a bar chart that illustrates the comparison of all the metrics between VGG-19, Efficient-Net, and the suggested model. Table 8 presents the results of this comparison.

\begin{table}[H]
\centering
\caption{Comparison of the suggested model with VGG-19, Efficient-Net, and VGG-16}
\renewcommand{\arraystretch}{1.3}

\begin{tabular}{|c|c|c|c|c|}
\hline
\textbf{S.No.} & \textbf{Metrics} & \textbf{Efficient-Net} & \textbf{VGG-19} & \textbf{Proposed model} \\
\hline
1 & \textbf{Accuracy (\%)} & 97.29 & 97.50 & \textbf{98.83} \\
\hline
2 & \textbf{Precision (\%)} & 94.36 & 95.25 & \textbf{97.70} \\
\hline
3 & \textbf{Recall (\%)} & 97.29 & 100 & \textbf{98.64} \\
\hline
4 & \textbf{F1-score (\%)} & 95.80 & 97.56 & \textbf{98.17} \\
\hline
\end{tabular}

\end{table}

\FloatBarrier
\begin{figure}[htbp]
\centering
\includegraphics[width=0.7\linewidth]{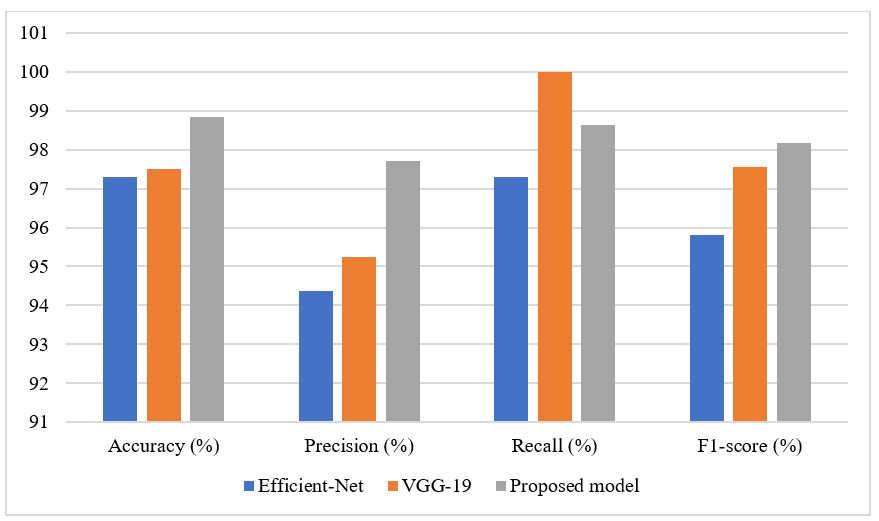}
\caption{Graph of comparison of the Suggested Model}
\end{figure}

6.4.2	Comparison of the suggested model with previous work
The comparative study of the suggested model with the results of earlier investigations is shown in Table 9. Compared to the model used in the earlier research, the suggested model achieves a higher level of accuracy, which is 98.83 percentage. This demonstrates that the model is effective. A comparison graph of the planned work with the work that has been done in the past is shown in Figure 19.

\begin{table}[H]
\centering
\caption{Comparison of the suggested model with previous work}
\renewcommand{\arraystretch}{1.3}

\begin{tabular}{|p{4.5cm}|c|p{4.5cm}|c|}
\hline
\textbf{Authors [Reference]} & \textbf{Year} & \textbf{Methods Used} & \textbf{Accuracy (\%)} \\
\hline
\textbf{Ihlen et al., (2019) [49]} & 2019 & Computer-based Infant Movement Assessment (CIMA) & 92.7 \\
\hline
\textbf{Goodlichh et al., (2020) [50]} & 2020 & Random Forest (RF) and SVM & 92 \\
\hline
\textbf{Bakheet et al., (2021) [51]} & 2021 & Random under-sampling boosting (RUS-Boost) & 84.6 \\
\hline
\textbf{Mutharika et al., (2021) [52]} & 2021 & LR & 83 \\
\hline
\textbf{Proposed Work} & 2024 & Proposed model & 98.83 \\
\hline
\end{tabular}

\end{table}

\FloatBarrier
\begin{figure}[htbp]
\centering
\includegraphics[width=0.7\linewidth]{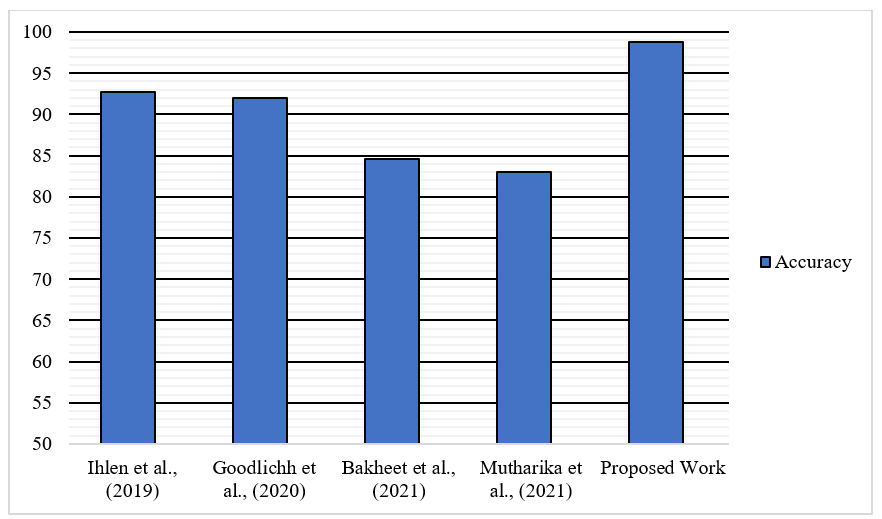}
\caption{Graph of comparison of the suggested model with other}
\end{figure}

\section{Conclusion}

CP is a group of permanent movement disorders that appear in early childhood, affecting posture, balance, and motor functions. Early and accurate detection of CP is crucial for timely intervention and management. This study presents a novel machine-learning model designed to detect CP disorder. The suggested model uses a large dataset consisting of brain MRI scans. Missing data is executed, features are normalized, and image quality is enhanced using various preprocessing approaches. The Bi-LSTM is a supervised learning algorithm used to train the ML model that combines feature extraction approaches. They used pre-train CNN models such as VGG-19 and Efficient-Net. Standard measures including F1-score, recall, accuracy, and precision are used to assess the model's performance. The experimental results show that classification accuracy of 97.29 percentage  when applying the Efficient-Net model, 97.50 percentage  when applying the VGG-19 model, and 98.83 percentage when the two models were combined with Bi-LSTM as a classifier. Thus, when the two models are combined, the extracted features are further improved. The accuracy of the model and dataset expansion to include a more varied patient group will be the primary goals of future studies.


\begin{thebibliography}{99}

\bibitem{ref1}
Australian Cerebral Palsy Register Group. ``Report of the Australian Cerebral Palsy Register, Birth Years 1993--2006.'' Cerebral Palsy Alliance Research Institute, Sydney (2013).

\bibitem{ref2}
Novak, I., Morgan, C., Adde, L., Blackman, J., Boyd, R.~N., Brunstrom-Hernandez, J., Cioni, G., et al.
``Early, accurate diagnosis and early intervention in cerebral palsy: advances in diagnosis and treatment.''
\textit{JAMA Pediatrics}, 171(9), 897--907 (2017).

\bibitem{ref3}
Te Velde, A., Tantsis, E., Novak, I., Badawi, N., Berry, J., Golland, P., Korkalainen, J., et al.
``Age of diagnosis, fidelity and acceptability of an early diagnosis clinic for cerebral palsy: a single site implementation study.''
\textit{Brain Sciences}, 11(8), 1074 (2021).

\bibitem{ref4}
Boychuck, Z., Andersen, J., Fehlings, D., Kirton, A., Oskoui, M., Shevell, M., Majnemer, A., et al.
``Current referral practices for diagnosis and intervention for children with cerebral palsy: a national environmental scan.''
\textit{The Journal of Pediatrics}, 216, 173--180 (2020).

\bibitem{ref5}
King, A.~R., Al Imam, M.~H., McIntyre, S., Morgan, C., Khandaker, G., Badawi, N., and Malhotra, A.
``Early diagnosis of cerebral palsy in low-and middle-income countries.''
\textit{Brain Sciences}, 12(5), 539 (2022).

\bibitem{ref6}
Yang, R., Zuo, H., Han, S., Zhang, X., and Zhang, Q.
``Computer-Aided Diagnosis of Children with Cerebral Palsy under Deep Learning CNN Image Segmentation Model Combined with 3D Cranial MRI.''
\textit{Journal of Healthcare Engineering}, 2021, 1822776 (2021).

\bibitem{ref7}
Clutterbuck, G., Auld, M., and Johnston, L.
``Active exercise interventions improve gross motor function of ambulant/semi-ambulant children with cerebral palsy: a systematic review.''
\textit{Disability and Rehabilitation}, 41(10), 1131--1151 (2019).

\bibitem{ref8}
Hakami, W.~S., Hundallah, K.~J., and Tabarki, B.~M.
``Metabolic and genetic disorders mimicking cerebral palsy.''
\textit{Neurosciences Journal}, 24(3), 155--163 (2019).

\bibitem{ref9}
Bosley, T.~M., Alorainy, I.~A., Oystreck, D.~T., Hellani, A.~M., Seidahmed, M.~Z., Osman, M.~E.~F., Sabry, M.~A., et al.
``Neurologic injury in isolated sulfite oxidase deficiency.''
\textit{Canadian Journal of Neurological Sciences}, 41(1), 42--48 (2014).

\bibitem{ref10}
Van Rappard, D.~F., Boelens, J.~J., and Wolf, N.~I.
``Metachromatic leukodystrophy: disease spectrum and approaches for treatment.''
\textit{Best Practice \& Research Clinical Endocrinology \& Metabolism}, 29(2), 261--273 (2015).

\bibitem{ref11}
Wiesinger, C., Eichler, F.~S., and Berger, J.
``The genetic landscape of X-linked adrenoleukodystrophy: inheritance, mutations, modifier genes, and diagnosis.''
\textit{The Application of Clinical Genetics}, 109--121 (2015).

\bibitem{ref12}
Cho, K.~H., Shim, S.~H., and Kim, M.
``Clinical, biochemical, and genetic aspects of Sjögren-Larsson syndrome.''
\textit{Clinical Genetics}, 93(4), 721--730 (2018).

\bibitem{ref13}
Suzuki-Muromoto, S., Wakusawa, K., Miyabayashi, T., Sato, R., Okubo, Y., Endo, W., Inui, T., et al.
``A case of new PCDH12 gene variants presented as dyskinetic cerebral palsy with epilepsy.''
\textit{Journal of Human Genetics}, 63(6), 749--753 (2018).

\bibitem{ref14}
McMichael, G., Haan, E., Gardner, A., Yap, T.~Y., Thompson, S., Ouvrier, R., Dale, R.~C., Gecz, J., and MacLennan, A.~H.
``NKX2-1 mutation in a family diagnosed with ataxic dyskinetic cerebral palsy.''
\textit{European Journal of Medical Genetics}, 56(9), 506--509 (2013).

\bibitem{ref15}
Pacheva, I.~H., Todorov, T., Ivanov, I., Tartova, D., Gaberova, K., Todorova, A., and Dimitrova, D.
``TSEN54 gene-related pontocerebellar hypoplasia type 2 could mimic dyskinetic cerebral palsy with severe psychomotor retardation.''
\textit{Frontiers in Pediatrics}, 6, 1 (2018).

\bibitem{ref16}
El-Hattab, A.~W., and Scaglia, F.
``Mitochondrial cytopathies.''
\textit{Cell Calcium}, 60(3), 199--206 (2016).

\bibitem{ref17}
Vafaee, A., Baghdadi, T., and Norouzzadeh, S.
``Cockayne syndrome misdiagnosed as cerebral palsy.''
\textit{Iranian Journal of Child Neurology}, 12(4), 162 (2018).

\bibitem{ref18}
Suzuki, R., Tanaka, A., Matsui, T., Gunji, T., Tohyama, J., Nairita, A., Nanba, E., and Ohno, K.
``Niemann-Pick Disease Type C presenting as a developmental coordination disorder.''
\textit{Case Reports in Pediatrics}, 2015, 807591 (2015).

\bibitem{ref19}
Parveen, S.
``Management and treatment for cerebral palsy in children.''
\textit{Indian Journal of Pharmacy Practice}, 11(2) (2018).

\bibitem{ref20}
Khan, H.~A., Jue, W., Mushtaq, M., and Mushtaq, M.~U.
``Brain tumor classification in MRI image using convolutional neural network.''
\textit{Mathematical Biosciences and Engineering} (2021).


\bibitem{ref21}
Horber, V., Grasshoff, U., Sellier, E., Arnaud, C., Krägeloh-Mann, I., and Himmelmann, K.
``The role of neuroimaging and genetic analysis in the diagnosis of children with cerebral palsy.''
\textit{Frontiers in Neurology}, 11, 628075 (2021).

\bibitem{ref22}
Mathew, S.~P., Dawe, J., Musselman, K.~E., Petrevska, M., Zariffa, J., Andrysek, J., and Biddiss, E.
``Measuring functional hand use in children with unilateral cerebral palsy using accelerometry and machine learning.''
\textit{Developmental Medicine \& Child Neurology} (2024).

\bibitem{ref23}
Sabater-Gárriz, Á., Gaya-Morey, F.~X., Buades-Rubio, J.~M., Manresa-Yee, C., Montoya, P., and Riquelme, I.
``Automated facial recognition system using deep learning for pain assessment in adults with cerebral palsy.''
\textit{Digital Health}, 10 (2024).

\bibitem{ref24}
Mohan, P.~P., and Ramkumar, G.
``Experimental evaluation of brain cerebral palsy disease prediction using artificial intelligence assisted learning methodology.''
In \textit{Proc. ICONSTEM}, IEEE (2024).

\bibitem{ref25}
Bertoncelli, C.~M., Bertoncelli, D., Bagui, S.~S., Bagui, S.~C., Costantini, S., and Solla, F.
``Identifying postural instability in children with cerebral palsy using a predictive model.''
\textit{Diagnostics}, 13(12), 2126 (2023).

\bibitem{ref26}
Gao, Q., Yao, S., Tian, Y., Zhang, C., Zhao, T., Wu, D., Yu, G., and Lu, H.
``Automating general movements assessment with quantitative deep learning.''
\textit{Nature Communications}, 14, 8294 (2023).

\bibitem{ref27}
Ramadhan, H.~H., Hussein, Q.~M., and Ahmed, M.~A.
``Cerebral palsy prediction using CNN depending on MRI images.''
\textit{J. Optoelectron Laser}, 41(8), 724--733 (2022).

\bibitem{ref28}
Li, D., Qu, J., Tian, Z., Mou, Z., Zhang, L., and Zhang, X.
``Knowledge-based recurrent neural network for TCM cerebral palsy diagnosis.''
\textit{Evidence-Based Complementary and Alternative Medicine}, 2022 (2022).

\bibitem{ref29}
Chahrogh, L.~K.
\textit{Hip Segmentation for Children with Cerebral Palsy Using Deep Learning}.
Southern Illinois University at Edwardsville (2022).

\bibitem{ref30}
Xue, B., Licis, A., Boyd, J., Hoyt, C.~R., and Ju, Y.-E.~S.
``Validation of actigraphy for sleep measurement in children with cerebral palsy.''
\textit{Sleep Medicine}, 90, 65--73 (2022).

\bibitem{ref31}
Zhu, M., Men, Q., Ho, E.~S.~L., Leung, H., and Shum, H.~P.~H.
``Interpreting deep learning based cerebral palsy prediction with channel attention.''
In \textit{IEEE EMBS BHI}, IEEE (2021).

\bibitem{ref32}
Pan, S.~J., and Yang, Q.
``A survey on transfer learning.''
\textit{IEEE Transactions on Knowledge and Data Engineering}, 22(10), 1345--1359 (2009).

\bibitem{ref33}
Tan, M., and Le, Q.
``EfficientNetV2: Smaller models and faster training.''
In \textit{International Conference on Machine Learning (ICML)}, pp.~10096--10106. PMLR (2021).

\bibitem{ref34}
Hassan, S.~M., Maji, A.~K., Jasiński, M., Leonowicz, Z., and Jasińska, E.
``Identification of plant-leaf diseases using CNN and transfer-learning approach.''
\textit{Electronics}, 10(12), 1388 (2021).

\bibitem{ref35}
Simonyan, K., and Zisserman, A.
``Very deep convolutional networks for large-scale image recognition.''
\textit{arXiv preprint arXiv:1409.1556} (2014).

\bibitem{ref36}
Liang, N., Yuan, L., Wen, X., Xu, H., and Wang, J.
``End-to-end retina image synthesis based on CGAN using class feature loss and improved retinal detail loss.''
\textit{IEEE Access}, 10, 83125--83137 (2022).

\bibitem{ref37}
Zhao, H., Li, H., Maurer-Stroh, S., and Cheng, L.
``Synthesizing retinal and neuronal images with generative adversarial nets.''
\textit{Medical Image Analysis}, 49, 14--26 (2018).

\bibitem{ref38}
Kamil, M.~Y.
``A deep learning framework to detect COVID-19 disease via chest X-ray and CT scan images.''
\textit{International Journal of Electrical and Computer Engineering}, 11(1) (2021).

\bibitem{ref39}
Nguyen, T.-H., Nguyen, T.-N., and Ngo, B.-V.
``A VGG-19 model with transfer learning and image segmentation for classification of tomato leaf disease.''
\textit{AgriEngineering}, 4(4), 871--887 (2022).

\bibitem{ref40}
Librenza-Garcia, D., Kotzian, B.~J., Yang, J., Mwangi, B., Cao, B., Lima, L.~N.~P., Bermudez, M.~B., Boeira, M.~V., Kapczinski, F., and Passos, I.~C.
``The impact of machine learning techniques in the study of bipolar disorder: A systematic review.''
\textit{Neuroscience and Biobehavioral Reviews}, 80, 538--554 (2017).

\bibitem{ref41}
Sarker, I.~H.
``Machine learning: Algorithms, real-world applications and research directions.''
\textit{SN Computer Science}, 2(3), 160 (2021).

\bibitem{ref42}
Dutta, N., Umashankar, S., Shankar, V.~K.~A., Padmanaban, S., Leonowicz, Z., and Wheeler, P.
``Centrifugal pump cavitation detection using machine learning algorithm technique.''
In \textit{IEEE International Conference on Environment and Electrical Engineering (EEEIC)}, pp.~1--6 (2018).

\bibitem{ref43}
Sundararajan, K., Garg, L., Srinivasan, K., Bashir, A.~K., Kaliappan, J., Ganapathy, G.~P., Selvaraj, S.~K., and Meena, T.
``A contemporary review on drought modeling using machine learning approaches.''
\textit{CMES: Computer Modeling in Engineering and Sciences}, 128(2), 447--487 (2021).

\bibitem{ref44}
Roshanfeker, B., Khadvi, S., and Rahmati, M.
``Sentiment analysis using deep learning on Persian text.''
In \textit{IEEE Iranian Conference on Electrical Engineering (ICEE)}, Tehran, Iran, pp.~1503--1508 (2017).

\bibitem{ref45}
Vaateekul, P., and Komsubha, T.
``A study of sentiment analysis using deep learning technique on Thai Twitter data.''
In \textit{International Joint Conference on Computer Science and Software Engineering (JCSSE)}, pp.~1--6 (2016).

\bibitem{ref46}
Li, D., and Qian, J.
``Text sentiment analysis based on long short-term memory.''
In \textit{IEEE International Conference on Computer Communication and the Internet (ICCCI)}, pp.~471--475 (2016).

\bibitem{ref47}
Zebin, T., Sperrin, M., Peek, N., and Casson, A.~J.
``Human activity recognition from inertial sensor time-series using batch normalized deep LSTM recurrent networks.''
In \textit{IEEE EMBC}, pp.~1--4 (2018).

\bibitem{ref48}
Elzayady, H., Sobhy, M., and Badran, K.
``Integrated bidirectional LSTM--CNN model for customers reviews classification.''
\textit{Journal of Engineering Science and Military Technologies}, 5(1), 14--20 (2021).

\bibitem{ref49}
Ihlen, E.~A.~F., Støen, R., Boswell, L., de Regnier, R.-A., Fjørtoft, T., Gaebler-Spira, D., Labori, C., et al.
``Machine learning of infant spontaneous movements for the early prediction of cerebral palsy: A multi-site cohort study.''
\textit{Journal of Clinical Medicine}, 9(1), 5 (2019).

\bibitem{ref50}
Goodlich, B.~I., Armstrong, E.~L., Horan, S.~A., Baque, E., Carty, C.~P., Ahmadi, M.~N., and Trost, S.~G.
``Machine learning to quantify habitual physical activity in children with cerebral palsy.''
\textit{Developmental Medicine \& Child Neurology}, 62(9), 1054--1060 (2020).

\bibitem{ref51}
Bakheet, D., Alotaibi, N., Konn, D., Vollmer, B., and Maharatna, K.
``Prediction of cerebral palsy in newborns with hypoxic-ischemic encephalopathy using multivariate EEG analysis and machine learning.''
\textit{IEEE Access}, 9, 137833--137846 (2021).

\bibitem{ref52}
Muthureka, K., Reddy, U.~S., and Janet, B.
``Implementation of multivariate logistic regression model for cerebral palsy identification using prenatal and perinatal risk factors.''
In \textit{IOP Conference Series: Materials Science and Engineering}, 1085(1), 012015 (2021).

\bibitem{ref53}
Gajbhiye, N., and Singh, K.~K.
``An optimized depression detection technique using behavioral analysis and machine learning.''
In \textit{Proc. 6th International Conference on Recent Advances in Information Technology (RAIT)}, pp.~1--5 (2025).

\bibitem{ref54}
Gajbhiye, N., and Singh, K.~K.
``Meta heuristic based optimized intelligent framework for kidney disease detection using deep learning.''
In \textit{4th OPJU International Technology Conference (OTCON)}, pp.~1--5 (2025).

\bibitem{ref55}
Singh, K.~K., Gajbhiye, N., and Mishra, G.~S.
``Exploring multi-stage deep convolutional neural network for medicinal plant disease diagnosis.''
In \textit{Proc. 6th International Conference on Deep Learning, Artificial Intelligence and Robotics (ICDLAIR)}, pp.~87--101 (2024).

\bibitem{ref56}
Gajbhiye, N., Singh, K.~K., and Mishra, G.~S.
``Enhancing crop disease detection systems with explainable AI techniques for deep learning models using spectral imaging.''
In \textit{Proc. 6th International Conference on Deep Learning, Artificial Intelligence and Robotics (ICDLAIR)}, pp.~110--126 (2024).


\end{thebibliography}
\end{document}